# An ordinal view of independence with application to plausible reasoning

Didier Dubois, Luis Fariñas del Cerro, Andreas Herzig and Henri Prade
Institut de Recherche en Informatique de Toulouse – C.N.R.S.
Université Paul Sabatier, 118 route de Narbonne
31062 TOULOUSE Cedex – FRANCE
Email: {dubois, farinas, herzig, prade}@irit.fr, Fax: (+33) 61.55.62.58

## Abstract

An ordinal view of independence is studied in the framework of possibility theory. We investigate three possible definitions of dependence, of increasing strength. One of them is the counterpart to the multiplication law in probability theory, and the two others are based on the notion of conditional possibility. These two have enough expressive power to support the whole possibility theory, and a complete axiomatization is provided for the strongest one. Moreover we show that weak independence is well-suited to the problems of belief change and plausible reasoning, especially to address the problem of blocking of property inheritance in exception-tolerant taxonomic reasoning.

## 0 INTRODUCTION

The notion of epistemic independence can be studied in the framework of reasoning under uncertainty. It can be derived naturally using conditioning: *"C is independent of A* iff one's opinion about C is not affected by the fact of knowing A." Dually, we say that *C depends on A* if it is not the case that C is independent of A. A synonymous expression for "C depends on A" is *"A is relevant for C"*. Traditionally, the formal basis of the dependence relation (also called relevance relation) is conditional probability: Given two events A and C and a probability measure Prob, *C is (probabilistically) independent* of A iff $Prob(C|A) = Prob(C)$.

In this paper we show that independence based on possibility theory (Zadeh 1978, Dubois and Prade 1988) has quite different properties. We present three definitions of independence and study their formal properties. We call them unrelatedness, strong and weak independence. We show that strong independence has enough expressive power to support the whole possibility theory, and we give a complete axiomatization not explicitly involving the underlying uncertainty theory. Then we apply these notions to the problems of belief revision and exception-tolerant reasoning.

Throughout the paper A, B, C, D and E stand for events belonging to a Boolean algebra of subsets of a set $\Omega$. We use the symbols: conjunction $\wedge$, disjunction $\vee$, and negation $\neg$. True and False are propositional constants denoting the true and false events respectively.

## 1 PROBABILISTIC INDEPENDENCE
### 1.1 The Multiplication Law

The standard definition of probabilistic independence is via the *Multiplication Law: A and C are probabilistically independent iff $Prob(A \wedge C) = Prob(A) * Prob(C)$*. A priori we find it more natural to define independence as: *C is probabilistically independent of A iff $Prob(C|A) = Prob(C)$*, relying thus on conditioning. In fact this conditioning-based notion of independence is equivalent to the multiplication law. It follows from the axioms of probability theory that independence defined in this way is a symmetric relation, and that it is not sensitive to negation. In other words:

(symmetry) If C is independent of A then A is independent of C.
(negation) If C is independent of A then C is independent of $\neg A$.

Symmetry justifies to say "A and C are independent" instead of "A is independent of C". Moreover we have

(truth) A and True are independent.

Note that there are no such simple properties governing the interplay of independence with conjunction and disjunction.

The above properties are not enough to completely characterize the notion of probabilistic independence. To get completeness we need two more axioms (Kolmogorov 1956), cited in (Fine 1973):

(prob $\vee$) If A and B are independent, A and C are independent and B and C are mutually exclusive, then A and B$\vee$C are independent.
(prob $\wedge$) If A and B are independent, C and D are independent, $B \geq D$ and $A \geq C$ then $A \wedge B \geq C \wedge D$.

Hence the axiomatisation of dependence involves not only logical truth, but also a qualitative probability relation "$\geq$". It follows that - at least for the simple definition via conditioning - we cannot study the formal properties of probabilistic dependence separately from the probabilistic framework.

The Multiplication Law has been criticized by several authors. R. von Mises (1964) has argued that the Multiplication Law could be fulfilled just because of "pure numerical accidents", although A and B are not intuitively independent in the sense of 'being separated' or 'not influencing each other'. He gives an example where Prob(A) = Prob(B) = Prob(C) = Prob(D) = 1/4, and A, B, C and D are pairwise mutually exclusive. Then he investigates whether A∨B and B∨C are independent. According to von Mises they are intuitively dependent (because they have B in common), whereas the Multiplication Law says that they are independent.

H. Reichenbach (1949) has argued that dependence and independence should be ternary rather than binary relations, where the third element in the relation is the evidence on which we declare that A and C are independent. From the ternary relation we can get back the binary one as follows: *A and C are independent iff there is some evidence E such that A and C are independent on evidence E.* We do not treat ternary relations in the sequel except in the last section. Nevertheless our analysis carries over to the ternary case. In the rest of the section we formally present two important principles that are not validated by the Multiplication Law.

### 1.2 Conjunction and (in)dependence

A formal objection to the multiplication Law has been given by J. M. Keynes (1921, cited in (Gärdenfors 1978)). According to Keynes, the following *conjunction criterion for dependence* (called conjunction criterion for relevance in (Gärdenfors 1978)) should be valid:

(CCD)   If C depends on A, and C depends on B then C depends on A∧B.

He notes that (CCD) is not validated by the Multiplication Law. Keynes proposes a stronger definition of probabilistic independence that does it: *C is independent of A iff there is no B such that A implies B and Prob(C|B) ≠ Prob(C).* R. Carnap (1950) has shown that this definition leads to a trivial notion of independence: It entails that C depends on A as soon as C and A are consistent.

Gärdenfors (1978) has suggested *a conjunction criterion for independence* dual to (CCD):

(CCI)   If C is independent of A, and C is independent of B then C is independent of A∧B.

is a natural principle that should be valid. Gärdenfors criticizes the Multiplication Law because it does not guarantee this principle, and investigates a series of stronger definitions of independence. He finally comes up with: *C is independent of A iff Prob(A) = 0, or Prob(C|A∧B) = Prob(C) for all B such that Prob(A∧B) > 0 and Prob(C|B) = Prob(C).* This definition validates (CCI).

Note that Gärdenfors' independence relation is non-symmetric. Here we show that in an ordinal setting where uncertainty is described by ordering the states of the world by their plausibility, we capture similar regularities in terms of disjunction, in a much simpler way.

## 2 POSSIBILITY THEORY

We introduce the notions of possibility measure and distribution and of conditional possibility.

### 2.1 Possibility Measures

Possibility measures allow to associate an uncertainty degree to the elements of the set of events. Following e.g., Dubois and Prade (1986), a function $\Pi$ from $2^\Omega$ into the real interval [0,1] is a *possibility measure* if it satisfies the following axioms:

$\Pi(\text{True}) > \Pi(\text{False})$  ;  $\Pi(A \vee B) = \max(\Pi(A), \Pi(B))$

Any totally ordered set can stand for the unit interval, that we keep here for the sake of simplicity. By convention $\Pi(\text{True}) = 1$, $\Pi(\text{False}) = 0$. $\Pi(A) = 1$ only means that A is possibly true, while $\Pi(A) = 0$ means that A is certainly false. Particularly, when $\Pi(A) = \Pi(\neg A) = 1$, it means total ignorance about the truth of A. The basic max-decomposability axiom is due to the purely ordinal setting. It can also be viewed as enforcing $\Pi(A \cup B)$ to its *lowest bound since for any reasonable confidence measure*, we should have "if A implies B then $\Pi(A) \leq \Pi(B)$". Note that $\Pi$ cannot be decomposable with respect to conjunction because $\Pi(A \wedge \neg A) = 0$ for all A, but $\Pi(A) = \Pi(\neg A) = 1$ is permitted. Moreover, $\max(\Pi(A), \Pi(\neg A)) = \Pi(\text{True})$, since $A \vee \neg A = \text{True}$. The quantity $N(A) = 1 - \Pi(\neg A)$ is called the necessity of A, and represents a level of certainty, or acceptance of A. Especially $N(A) > 0$ means that A is accepted, while ¬A is not (since $N(A) > 0$ entails $N(\neg A) = 0$ in the absence of inconsistency). And we have the reasonable axiom of acceptance saying that if A is accepted and so is B, then A∧B is accepted too, since $N(A \wedge B) = \min(N(A), N(B))$ holds. Note that the fact that A is not accepted ($N(A) = 0$) does not entail that it is rejected (which is expressed by $N(\neg A) > 0$).

In the finite case a possibility measure can be represented by a possibility distribution $\pi$ on $\Omega$ the set of interpretations of the language. Namely $\Pi(A) = \sup\{\pi(\omega) \mid \omega \models A\}$. $\pi$ encodes a complete partial ordering of interpretations, with the intended meaning that if $\pi(\omega) > \pi(\omega')$, $\omega$ is a more plausible description of the current situation than $\omega'$. Reasoning in the setting of possibility theory comes down to assume that the current situation is always one of the most plausible ones. Accepting A ($N(A) > 0$) means that $\Pi(A) > \Pi(\neg A)$, i.e., that "normally A should be true". This way of modelling uncertainty is in full accordance with Shoham (1988)'s view of preferential logic, Lehmann (1989)'s notion of ranked models, also encoded in Pearl (1990)'s system Z. See (Benferhat et al. 1992). Especially $N(A) > 0$ can be written True $\vdash$ A in the terminology of conditional assertions in Lehmann's rational nonmonotonic setting. True $\vdash$ A means "A is plausibly true" (unconditionally).

Every possibility measure induces a relation "≥" defined by A ≥ B if and only if $\Pi(A) \geq \Pi(B)$. We call ≥ a relation *agreeing strictly* with $\Pi$. A ≥ B is read "A is at least as possible as B". This relation is called *qualitative possibility ordering* and satisfies the following conditions:

(non triviality)   True > False
(tautology)        True ≥ A
(transitivity)     if A ≥ B and B ≥ C then A ≥ C
(disjunctiveness)  A∨C ≤ A  or  A∨C ≤ C.
(dominance)        If A implies B then A ≤ B

**Remark.** Often, (disjunctiveness) is replaced by the two axioms of connectedness (A ≥ B or B ≥ A) and disjunctive stability (if A ≥ B then A∨C ≥ B∨C)

That these conditions are sufficient follows from the following formal relation between possibility theory and qualitative possibility orderings (Dubois 1986): The only functions mapping events into [0,1] which strictly agree with qualitative possibility orderings are possibility measures, and a strictly agreeing possibility measure always exists. In our presentation of possibility theory, this result can be expressed as follows:

**Theorem** (soundness and completeness of qualitative possibility orderings). Let $\Pi$ be any measure on $\Omega$, and ≥ any binary relation on the subsets of $\Omega$. Then $\Pi$ is a possibility measure iff ≥ is a qualitative possibility ordering.

Conditional possibility relations first appear in Lewis (1973)'s logics of counterfactuals whose semantics are systems of spheres. But as indicated in Dubois and Prade (1991) a system of spheres is equivalent to a possibility distribution. Formal links between possibility theory and conditional logics are studied in Fariñas del Cerro and Herzig (1991). Lastly, the dual qualitative certainty relation, equivalent to necessity measures (A $\geq_N$ C iff ¬C ≥ ¬A) is closely related to expectation-orderings of Gärdenfors and Makinson (1994). The characteristic axiom of $\geq_N$ is

(conjunctiveness)     A∧C $\geq_N$ A  or  A∧C $\geq_N$ C.

### 2.2 Conditional Possibility

Following Hisdal (1978) and Dubois and Prade (1986, 1992) the *conditional possibility* $\Pi(C|A)$ is defined as the maximal solution of the equation $\Pi(A∧C) = \min(\Pi(C|A), \Pi(A))$. This definition is clearly inspired from Bayes' Rule, where min corresponds to the product. The choice of the maximal solution is due to the *principle of minimal specificity* which urges to select the least committed possibility measures, i.e. the one which allows each event to have the greatest possibility level:

$\Pi(C|A) = 1$          if $\Pi(A) = \Pi(A∧C)$
$\Pi(C|A) = \Pi(A∧C)$   if $\Pi(A) > \Pi(A∧C)$

**Facts.**
1. If $\Pi(A) = 0$ then $\Pi(C|A) = 1$
2. If $\Pi(A∧C) = 1$ then $\Pi(C|A) = 1$
3. If $\Pi(A) > 0$ and $\Pi(C) = 0$ then $\Pi(C|A) = 0$
4. $\Pi(C|¬C) = 1$ iff $\Pi(¬C) = 0$
5. $\Pi(C|¬C) = 0$ iff $\Pi(¬C) > 0$

Some of these facts deserve some comments. Fact 1 suggests that nothing is sure when assuming that a certainly false proposition is true (since in this case anything and its contrary is plausible). This leads to the convention $\Pi(False|False) = 1$ which does not agree with the non-triviality axiom (it is not compulsory anyway). Fact 2 says that if A and C are fully consistent, assuming A true keeps C possibly true. Fact 3 says that a certainly false proposition remains false via conditioning by a non-certainly false proposition. However if $\Pi(A) = 0$ then the conditional possibility again disagrees with the non-triviality axiom.

In the next sections we present three different ordinal definitions of (in)dependence. Two of them are based on the notion of conditional possibility. We show that these two can express qualitative possibility, and that complete axiomatizations is given for one of them. We conjecture that this is not possible for the third one, originally due to Zadeh. In all three cases, a necessary condition for the independence of A and C will be that the conjunction A∧C can be interpreted truth-functionally, in the sense that $\Pi(A∧C) = \min(\Pi(A), \Pi(C))$ for these particular events.

The conditional necessity is $N(¬C|A) = 1 - \Pi(C|A)$, defined by duality. Note that $N(C|A) = N(¬A∨C)$ if $N(C|A) > 0$. The following property will be used at length in the sequel:

$$N(C|A) > 0 \text{ iff } \Pi(A∧C) > \Pi(A∧¬C)$$

$N(C|A) > 0$ means that C is accepted as true when A is assumed to be true. It corresponds to the conditional assertion A ⊢ C in the sense of Lehmann's rational inference, and can be viewed as the (nonmonotonic) plausible entailment of C from A in the presence of an ordering of interpretations. The above clearly show that A ⊢ C means A∧C is more plausible than A∧¬C (or equivalently $\Pi(A∧C) > \Pi(A∧¬C)$ in terms of possibility measure).

## 3 (UN)RELATEDNESS

Zadeh (1978) has introduced a symmetric definition of independence called "non-interactivity" between possibilistic variables that is not based on conditional possibilities. This notion has also been studied by Nahmias (1978) for events, under the name "unrelatedness".

(**Def** $≈_Z$)  *A and C are related propositions in Zadeh's sense* (denoted by A $≈_Z$ C) iff $\Pi(A∧C) ≠ \min(\Pi(A), \Pi(C))$.

It is interesting to characterize the constraints induced by unrelatedness on the ordering of interpretations A∧C, ¬A∧C, A∧¬C, ¬A∧¬C respectively.

**Proposition.** A and C are unrelated if and only if $\Pi(A∧C) \geq \min(\Pi(A∧¬C), \Pi(¬A∧C))$.
*Proof.* $\Pi(A∧C) = \min(\max(\Pi(A∧C), \Pi(A∧¬C)), \max(\Pi(A∧C), \Pi(¬A∧C)))$. Clearly as soon as $\Pi(A∧C) \geq \Pi(A∧¬C)$, unrelatedness holds. And the same when $\Pi(A∧C) \geq \Pi(¬A∧C)$. However if $\Pi(A∧C) < \Pi(A∧¬C)$ and $\Pi(A \wedge C) < \Pi(\neg A \wedge C)$, then $\Pi(A) = \Pi(A \wedge \neg C)$ and $\Pi(C) = \Pi(\neg A \wedge C)$, and A and C are related.

Clearly, A and C are related implies that $A \wedge C$ is an implausible situation (since in any case it holds $\Pi(A \wedge C) \leq \min(\Pi(A), \Pi(C))$ ), i.e., A and C are (more or less) mutually exclusive:

**Corollary.** A and C are related if and only if $N(\neg C|A) > 0$ and $N(\neg A|C) > 0$.

On the contrary when A and C are unrelated the two propositions are totally allowed to be true together. Zadeh's independence is an extension of the logical notion of consistency. This notion is not very demanding. Moreover this notion is local in the sense that it is sensitive to negation: if A and C are unrelated, it does not say anything about the other literals $\neg A$ and C, C and $\neg A$, $\neg C$ and $\neg A$. Other properties are as follows.

**Facts.**
1. $A \approx_Z C$ iff $C \approx_Z A$
2. If $A \approx_Z B \vee C$ then $A \approx_Z B$ or $A \approx_Z C$
3. If $A \vee B \approx_Z C$ then $A \approx_Z C$ or $B \approx_Z C$
   (due to symmetry)
4. If $A \approx_Z C$ and $B \approx_Z C$ then $A \vee B \approx_Z C$
5. If $A \approx_Z B$ and $A \approx_Z C$ then $A \approx_Z B \vee C$
   (due to symmetry)
6. False $\not\approx_Z A$ (where $\not\approx_Z$ means "not($\approx_Z$ )")
7. True $\not\approx_Z A$   ;   8. $A \not\approx_Z A$
9. $A \not\approx_Z \neg A$  iff  $\Pi(A) = 0$ or $\Pi(\neg A) = 0$ ;  10. $A \vee C \not\approx_Z A$

Facts 2 and 3 are disjunction-oriented. However, none of the two conjunction criteria (CCD) and (CCI) are valid with unrelatedness. Note also that facts 8 and 10 is certainly a strange property for an independence relation. There seems to be no way to express $\Pi$ by means of $\approx_Z$, the reason being that we cannot express $\Pi(A) = 1$. Therefore, we conjecture that (just as for probabilistic independence) $\approx_Z$ cannot be axiomatized alone.

## 4 STRONG INDEPENDENCE

It is tempting to define dependence in possibility theory in a way similar to probability theory, namely to define C as independent of A when the conditional measure of C given A is equal to the unconditional measure of C. Here we have two uncertainty functions $\Pi$ and N. Hence we can define independence as $\Pi(C|A) = \Pi(C)$ or $N(C|A) = N(C)$. Notice that $N(C|A) = N(C)$ is equivalent to $\Pi(\neg C|A) = \Pi(\neg C)$. In (Fariñas and Herzig 1994a) the independence relation defined by $\Pi(C|A) = \Pi(C)$ is studied. A complete axiomatisation has been given.

Note that if $\Pi(C|A) = \Pi(C) < 1$ then we are in the situation where C is plausibly rejected (since $\Pi(\neg C) = 1 > \Pi(C)$). Hence the meaning of $\Pi(C|A) = \Pi(C) < 1$ is that when A is assumed to be true, it does not affect the plausible rejection of C. This expresses the negative statement that accepting $\neg C$ is independent of A. It suggests to use $N(C|A) = N(C)$ in order to express a positive statement. Note that we also have

$\Pi(C|A) = \Pi(C) < 1$ implies $\Pi(\neg C|A) = \Pi(\neg C) = 1$

but not the converse. Hence $\Pi(\neg C|A) = \Pi(\neg C) = 1$ is a very weak statement saying that not accepting C (i.e. $N(C) = 0$) is not questioned by fact A. In particular, $\Pi(C|A) = \Pi(C) = \Pi(\neg C|A) = \Pi(\neg C) = 1$ (which is never met in the probabilistic case), means that in the presence of A, C, which was originally ignored, is still ignored. In this paper we shall restrict to independence of *accepted* propositions with respect to other propositions; independence of *ignored* propositions turns out to be a very distinct issue, as suggested by the following result:

**Proposition 4.1.** $N(C|A) = N(C)$ iff either (i) $1 = \max(\Pi(\neg A \wedge \neg C), \Pi(A \wedge \neg C))$, and $\Pi(A \wedge \neg C) \geq \Pi(A \wedge C)$, or (ii) $\Pi(A \wedge C) > \Pi(A \wedge \neg C) \geq \Pi(\neg A \wedge \neg C)$

Moreover, (i) is equivalent to $N(C|A) = N(C) = 0$, and (ii) is equivalent to $N(C|A) = N(C) > 0$. Note that the two situations (i) and (ii) correspond to (almost) reversed orderings of interpretations. We give the following definition of the *strong independence relation*:

**(Def $\neq$>)** C is strongly independent of A (denoted by $A \neq> C$) iff $N(C|A) = N(C) > 0$.

Note that $A \neq> C$ indicates that in the context where A is true C is accepted. Due to what we said above, C is strongly independent of A iff $\Pi(\neg C|A) = \Pi(\neg C) < 1$.

In the next theorem we characterize a dependence relation $\approx> = \text{not}(\neq>)$ without using conditional necessities.

**Theorem (construction of $\approx>$ from $\Pi$).** Let $\Pi$ be a possibility measure, and let $\approx>$ be defined from its dual N through (Def $\neq$>).
1. $A \neq> C$  iff $\Pi(A) > \Pi(\neg C) = \Pi(A \wedge \neg C)$
2. $A \approx> C$  iff $\Pi(A) \leq \Pi(\neg C)$ or $\Pi(\neg C) > \Pi(A \wedge \neg C)$

*Proof.* Follows directly from Proposition 4.1.

**Corollary.** Let $\Pi$ be a possibility measure, and let $\approx>$ be defined from $\Pi$ through (Def $\neq$>).
1. $A \neq> C$ iff $\Pi(A \wedge \neg C) = \min(\Pi(A), \Pi(\neg C))$ and $\Pi(\neg C) < \Pi(A)$.
2. $A \approx> C$ iff $\Pi(A \wedge \neg C) \neq \min(\Pi(A), \Pi(\neg C))$ or $\Pi(\neg C) \geq \Pi(A)$.
3. If $A \neq> C$ then $\Pi(A \wedge C) = \min(\Pi(A), \Pi(C))$.
4. If $\Pi(\neg C) \geq \Pi(A)$ then $A \approx> C$.

**Facts.**
1. If  $A \approx> B \wedge C$  then  $A \approx> B$  or  $A \approx> C$
2. If  $A \vee B \approx> C$  then  $A \approx> C$  or  $B \approx> C$
3. If  $A \approx> C$  and  $B \approx> C$  then  $A \vee B \approx> C$
4. If  $A \approx> B$  and  $A \approx> C$  then  $A \approx> B \wedge C$
5. False $\approx> C$   ;   6. True $\neq> C$  iff  $N(C) > 0$
7. $A \approx>$ False   ;   8. $A \neq>$ True  iff  $\Pi(A) > 0$
9. $A \wedge B \approx> \neg B \wedge C$   10. If A implies $\neg C$ then $A \approx> C$
11. $A \vee C \neq> \neg C$   iff   $\Pi(A) > \Pi(C)$
12. $A \neq> A$ iff $N(A) = 1$  ;  13. If $\Pi(A) = 0$ then $A \approx> C$
14. If $\Pi(C) = 1$ then $A \approx> \neg C$
15. $A \approx> C$ or $\neg C \approx> \neg A$

Let us comment on these facts. Facts 2 and 3 are similar to the (CCI) and (CCD) axioms except that disjunction is used instead of conjunction. Facts 1 and 4 are also similar but the conjunction of influenced facts is considered instead of influencing facts. Fact 5 means that assuming a contradiction holds destroys all previously plausible propositions. On the contrary tautologies never affect the plausibility of already plausible propositions (Fact 6). Fact 7 is simply due to the impossibility to assert false propositions. Fact 8 says that we can only assert a tautology is plausible when taking for granted an impossible proposition. Fact 9 and 10 express equivalent properties. Namely if A implies that C is false then when learning that A is true affects our opinion about C when C was previously supposed to be plausible. Fact 11 shows that the possibilistic ordering can be translated in terms of strong independence. Fact 12 claims that the only case when the truth of A is independent of itself is when A is a tautology. Fact 13 is a more general statement than fact 5. Fact 14 holds because it cannot be the case that $\Pi(A) > \Pi(C)$. Similarly the reason for fact 15 is that $\Pi(A) > \Pi(\neg C)$ cannot go along with $\Pi(\neg C) > \Pi(A)$.

Clearly, probabilistic dependence and possibilistic dependence are quite different concepts. Probabilistic properties such as symmetry ("If B depends on A then A depends on B") or transparency w.r.t. negation ("If B depends on A then B depends on $\neg A$") do not hold in the possibilistic case. In other words, $A \not\approx> B$ neither implies $B \not\approx> A$ nor $A \not\approx> \neg B$. On the other hand, possibilistic dependence has some "nice" regularities such as 1., 2., 3., 4., none of which holds in the probabilistic case. These regularities are quite close to the criteria (CCD) and (CCI).

Concerning the expressivity of the dependence relation it is interesting to observe that it possesses the same expressivity as possibility theory itself. This follows from the next result.

**Theorem** (construction of $\Pi$ from $\not\approx>$). Let $\Pi$ be a possibility measure, and let $\not\approx>$ be defined from $\Pi$.
1. $\Pi(A) > \Pi(C)$ iff $A \vee C \not\approx> \neg C$.
2. $\Pi(A) \geq \Pi(C)$ iff $A \vee C \approx> \neg A$.

*Proof.* By previous fact 11.

The theorem can be read as follows: C is strictly less possible than A if and only if learning that $A \vee C$ is true does not change my rejection of C. The theorem should not be surprising since the meaning of independence is to enforce constraints on the ordering between interpretations as shown in Proposition 4.1. It turns out that such constraints are enough to identify a single ordering, i.e. a comparative necessity relation.

Thus we are able to express qualitative possibility by means of strong independence. In a trivial manner, this correspondence enables us to obtain an axiomatization of the (in)dependence relation by translating the qualitative counterpart of possibility theory. Note that this is in contrast with probability theory: There, the independence relation cannot completely capture qualitative probability (which in turn determines the probability measure). Here we give a simpler *axiomatization* of $\approx>$:

($\approx>$ 1)  True $\not\approx>$ True
($\approx>$ 2)  A $\approx>$ False

($\approx>$ 3)  If $A \vee B \approx> \neg B$ and $B \vee C \approx> \neg C$ then $A \vee C \approx> \neg C$
($\approx>$ 4)  A $\approx>$ $\neg A$
($\approx>$ 5)  If $A \approx> B \wedge C$ then $A \approx> B$ or $A \approx> C$

**Theorem** (soundness and completeness of the axiomatics of $\approx>$ w.r.t. possibility theory). Let $\approx>$ be a relation on events, and $\Pi$ a mapping from the set of events to [0,1] such that $A \not\approx> C$ iff $N(C|A) = N(C) > 0$. Then $\approx>$ is a dependence relation iff N is a necessity measure.

*Proof.* From the right to the left, it is sufficient to prove that the above axioms (rewritten as qualitative necessities) are valid. Then we can use the soundness of qualitative necessity orderings w.r.t. possibility theory. From the left to the right, we prove that the axioms for qualitative necessity orderings are derivable from the above axiomatics (and then use the completeness of qualitative necessity orderings w.r.t. possibility theory). Using the previous theorem in terms of necessities, namely $N(A) > N(C)$ iff $\neg A \vee \neg C \not\approx> A$; $N(A) \geq N(C)$ iff $\neg A \vee \neg C \approx> C$ we express qualitative necessities with $\approx>$:

1. **(non triviality)** True $>_N$ False becomes
      $\neg False \vee \neg True \not\approx>$ True.
It is equivalent to True $\not\approx>$ True which is an instance of ($\approx>$ 1).

2. **(transitivity)** if $A \geq_N B$ and $B \geq_N C$ then $A \geq_N C$ becomes: If $\neg A \vee \neg B \approx> B$ and $\neg B \vee \neg C \approx> C$
      then $\neg A \vee \neg C \approx> C$ which is ($\approx>$ 3).

3. **(tautology)** $A \leq_N$ True becomes $\neg A \vee \neg True \approx>$ True which is nothing else but ($\approx>$ 2).

4. **(conjunctiveness)** $A \wedge C \geq_N A$ or $A \wedge C \geq_N C$ becomes $\neg(A \wedge C) \vee \neg A \approx> A$ or $\neg(A \wedge C) \vee \neg C \approx> C$, hence $\neg A \vee \neg C \approx> A$ or $\neg A \vee \neg C \approx> C$. The latter can be proved combining $\neg A \vee \neg C \approx> A \wedge C$ which is an instance of ($\approx>$ 4), and: If $\neg A \vee \neg C \approx> A \wedge C$ then $\neg A \vee \neg C \approx> A$ or $\neg A \vee \neg C \approx> C$ which is an instance of ($\approx>$ 5).

5. **(dominance)** can be replaced by
(equivalence)    If $A \leftrightarrow C$ then  $A \leq_N C$ and
(monotony)       $A \geq_N A \wedge C$.
The latter is translated to $\neg A \vee \neg(A \wedge C) \approx> A \wedge C$, which is an instance of ($\approx>$ 4). Hence what remains to prove is
      If $A \leftrightarrow C$ then  $\neg A \vee \neg C \approx> C$.
Now from $A \leftrightarrow C$ we get $\neg C \leftrightarrow \neg A \vee \neg C$. From the latter we get ($\neg C \approx> C$ iff $\neg A \vee \neg C \approx> C$). Then $\neg A \vee \neg C \approx> C$ follows from ($\approx>$ 4).

**Remark.** It is important to note that $\approx>$ is quite close to a qualitative possibility ordering: Replacing $A \approx> C$ by $\Pi(A) \leq \Pi(\neg C)$ all our principles are possibilistically valid. In particular (connectedness) can be deduced from the axioms: From ($\approx>$ 4) and ($\approx>$ 5) we can get $A \vee C \approx> \neg A$ or $A \vee C \approx> \neg C$ (see above).

The other way round, the only (qualitative) axiom for $\leq_N$ that apparently does not follow from the above axioms is that of transitivity. As on the other hand we know by the above Corollary that $A \not\approx> C$ implies $\Pi(A) > \Pi(\neg C)$, we obtain that for a given $\Pi$, $\not\approx>$ is a fragment of the corresponding strict possibility ordering. This fragment is closed under all the axioms of possibility theory except that of transitivity.

## 5  WEAK INDEPENDENCE

The notion of strong independence may be felt too strong because what we may wish to express is a more qualitative notion of independence. Now, strong independence requires that not only C remains more plausible than $\neg C$ when A is known to be true, but its level of acceptance should not be altered. This last requirement forces the inequality $\Pi(A \wedge \neg C) \geq \Pi(\neg A \wedge \neg C)$ which implies that in the context where C would be false, it is forbidden to conclude that $\neg A$ should be accepted (see Fact 15 of the previous section). Hence we have the property

C is strongly independent of A if and only if $N(C|A) > 0$ and $\neg(N(\neg A|\neg C) > 0)$.

A milder notion of independence is that if C is accepted unconditionally, then if A is true, C remains accepted; then we do away with any commitment in the case when C would turn out to be false. Hence the following definition:

(**Def** $\neq>_w$) *C is weakly independent of A*
(denoted $A \neq>_w C$) iff $N(C|A) > 0$ and $N(C) > 0$.

**Proposition 5.1.** $A \neq>_w C$ iff $\Pi(A \wedge C) > \Pi(A \wedge \neg C)$ and $\max(\Pi(C \wedge A), \Pi(C \wedge \neg A)) > \Pi(\neg A \wedge \neg C)$.
*Proof.* Indeed $A \neq>_w C$ is equivalent to $\Pi(A \wedge C) > \Pi(A \wedge \neg C)$, $\Pi(C) = 1 = \max(\Pi(C \wedge A), \Pi(C \wedge \neg A)) > \Pi(A \wedge \neg C)$ and $\max(\Pi(C \wedge A), \Pi(C \wedge \neg A)) > \Pi(\neg A \wedge \neg C)$, the first of which is redundant.    Q.E.D.

**Proposition 5.2.** $A \neq> C$ iff $A \neq>_w C$ and $\Pi(A \wedge \neg C) = \Pi(\neg C)$. (Obvious using Proposition 4.1.)

**Proposition 5.3.** $A \neq_w C$ implies $A \neq>_Z C$.
*Proof.* Obvious since $\Pi(A \wedge C) > \Pi(A \wedge \neg C)$, and then $\Pi(A) = \Pi(A \wedge C) \leq \Pi(C)$.

However it *is not true* that, as for strong independence

$A \neq>_w C$ implies $\Pi(A \wedge \neg C) = \min(\Pi(A), \Pi(\neg C))$

since weak independence does not involve $\Pi(A \wedge \neg C)$.

It can be checked that weak independence satisfies Facts 1, 2, etc. of the previous section except for a few ones, namely Fact 12, which becomes $A \neq>_w A$ iff $N(A) > 0$, and Fact 15. The latter is not surprising since weak independence is meant to let the relationship between $N(A|B) > 0$ and $N(\neg B|\neg A) > 0$ loose. Hence it is possible to have $A \neq>_w C$ and $\neg C \neq>_w \neg A$. This occurs precisely when $\Pi(\neg A \wedge C) > \max(\Pi(A \wedge C), \Pi(\neg A \wedge \neg C))$ and $\min(\Pi(A \wedge C), \Pi(\neg A \wedge \neg C)) > \Pi(A \wedge \neg C)$. Besides, weak independence satisfies stronger forms of Facts 3 and 4 :

3'.     if $A \approx>_w C$ or $B \approx>_w C$ then $A \vee B \approx>_w C$
4'.     if $A \approx>_w B$ or $A \approx>_w C$ then $A \approx>_w B \wedge C$

Lastly we have the following remarkable property

$$\forall A, C, \; A \vee \neg C \neq>_w C \text{ iff } A \vee \neg C \neq> C$$

Indeed if $A \vee \neg C \neq>_w C$, then $N(C|A \vee \neg C) > 0$ is such that $N(C|A \vee \neg C) = N((\neg A \wedge C) \vee C) = N(C)$.

Hence when C is weakly independent of A then it is also strongly independent of A as soon as $\neg C \vdash A$. As a consequence, it is easy to see that the theorem that constructs a possibility measure from the independence relation also holds when we change the strong independence into the weak independence. In fact only the part of the strong independence relation that is equivalent to weak independence is useful to recover the underlying possibility measure. However if $\neg C \vdash A$ does not hold, $A \neq>_w C$ does not enforce an inequality between $\Pi(C)$ and $\Pi(\neg A)$ generally. Finally the six axioms that characterize strong independence with respect to possibilistic semantics also hold for weak independence, but more axioms are necessary to completely characterize weak independence.

Let us show how weak independence can be related to the framework of belief revision (Gärdenfors, 1988). A central problem for the theory of belief revision is what is meant by a minimal change of a state of belief. As pointed out in Gärdenfors (1990), "the criteria of minimality that have been used [in the models for belief change] have been based on almost exclusively logical considerations. However, there are a number of non-logical factors that should be important when characterizing a process of belief revision". Gärdenfors focuses the notion of dependence (he uses the synonymous term 'relevance') and proposes the following criterion: *If a belief state K is revised by a sentence A, then all sentences in K that are independent of the validity of A should be retained in the revised state of belief.*

This seems to be a very natural requirement for belief revision operations, as well as a useful tool when it comes to implement belief change operations. As noted by Gärdenfors, "a criterion of this kind cannot be given a technical formulation in a model based on belief sets built up from sentences in a simple propositional language because the notion of relevance is not available in such a language." However the above criterion does make sense in the ordinal setting of possibility theory.

We suppose given a theory K and an AGM revision operation * (Gärdenfors, 1988). K*A represents the result of revising K by A. According to Gärdenfors and Makinson's characterization theorem, K and * can be represented equivalently by an epistemic entrenchment ordering, which in turn is nothing else than a qualitative necessity ordering. It can be proved that in terms of possibility theory the fact that C belongs to K*A is equivalent to having $N(C|A) > 0$ (Dubois and Prade, 1992); moreover C belongs to K is equivalent to $N(C) > 0$. If we translate the definition of the weak independence relation $\neq>_w$ in terms of revision we get

$$A \neq>_w C \text{ iff } C \in K \text{ and } C \in K*A$$

which is exactly Gärdenfors' above requirement for revision-based independence. Clearly, a companion definition of a dependence relation $\approx>^-$ associated to a given qualitative necessity ordering can be defined via the following condition from a given AGM contraction operation (-):

(**Cond** $\approx>^-$) $A \approx>^- C$ iff $C \in K$ and $C \notin K\text{-}A$
iff $N(C) > 0$ and $N(A) \geq N(A \vee C)$.

This alternative notion is studied in (Farinas and Herzig, 1994b). The comparative analysis of revision-based and contraction-based notions of independence is beyond the scope of this paper.

## 6  COMPARATIVE DISCUSSION

We have analysed three notions of (in)dependence that can be defined in possibility theory. A common feature to all of them is that the independence of A and C requires that the conjunction of A and C is interpreted truth-functionally. In other words, we have

$A \neq > C$ implies $A \neq >_W C$ ;   $A \neq >_W C$ implies $A \neq_Z C$

Hence, all notions of independence share the property $\Pi(A \wedge C) = \min(\Pi(A), \Pi(C))$. Moreover, we have shown that

$A \neq > C$ iff $A \neq_Z \neg C$ and $\Pi(\neg C) < \Pi(A)$
$A \neq > C$ iff $A \neq >_W C$ and $\Pi(A \wedge \neg C) = \Pi(\neg C)$

We now examine the validity of Keynes-Gärdenfors criteria of Section 1 in the ordinal setting of possibility theory. Namely the following requirements:

(CCD)   If $A \approx > C$ and $B \approx > C$ then $A \wedge B \approx > C$
(CCI)    If $A \neq > C$ and $B \neq > C$ then $A \wedge B \neq > C$

Also consider symmetric counterparts of CCD and CCI:

(CCD-r)  If $A \approx > B$ and $A \approx > C$ then $A \approx > B \wedge C$
(CCI-r)   If $A \neq > B$ and $A \neq > C$ then $A \neq > B \wedge C$

and the corresponding properties changing conjunction into disjunction (DCD, DCI, etc).

(DCI)    If $A \neq > C$ and $B \neq > C$ then $A \vee B \neq > C$
(DCI-r)  If $A \neq > B$ and $A \neq > C$ then $A \neq > B \vee C$
(DCD)   If $A \approx > C$ and $B \approx > C$ then $A \vee B \approx > C$
(DCD-r) If $A \approx > B$ and $A \approx > C$ then $A \approx > B \vee C$

First the relatedness property of Zadeh $\neq_Z$ satisfies the four above criteria concerning disjunctions. (CCD-r), (CCI-r), (DCI) (DCD) hold for strong and weak independence. The weak independence has the following stronger property:

$A \neq >_W B \wedge C$ iff $A \neq >_W B$ and $A \neq >_W C$
$A \vee B \neq >_W C$ iff $A \neq >_W C$ and $B \neq >_W C$

that is (DCI) and (CCI-r) with equivalence, due to Facts 3' and 4' of Section 5. This is natural if weak independence is considered in terms of belief revision: if we continue to accept $B \wedge C$ upon learning A we should continue to accept C and B as well.

We could have introduced as well a ternary dependence relation "B and C are independent, given A", as studied by Gärdenfors (1978, 1990) and Pearl (1988). For reasons of simplicity we have restricted our analysis to binary dependence relations here, but it is clear that a ternary relation is certainly the most general one. This will be subject of further investigations.

## 7  APPLICATION TO EXCEPTION-TOLERANT REASONING

Possibility theory is a natural framework for handling nonmonotonic reasoning problems, because it embeds what Lehmann calls rational inference (see Benferhat et al., 1992). Given a set of rules modelled by pairs of propositional formulas, it is possible to rank-order these exception-tainted rules in terms of their relative specificity. This ranking of rules generates an ordering of interpretations that can be encoded as a possibility distribution.

The algorithm for ranking rules (or interpretations) has been proposed by Lehmann, and also in a different form by Pearl. Benferhat et al. (1992) have shown that this ordering can be retrieved by means of the least specific possibility distribution that is consistent with the rules. Namely let K be a conditional knowledge base where rules are of the form $A_i \vdash B_i$ (read if $A_i$ is true, $B_i$ is plausibly true). Each rule is interpreted as the constraint $N(B_i | A_i) > 0$ or equivalently $\Pi(A_i \wedge B_i) > \Pi(A_i \wedge \neg B_i)$. Then the ranking of the interpretation obtained by considering the maximal element of the set $\{\pi, \Pi(A_i \wedge B_i) > \Pi(A_i \wedge \neg B_i), \forall i=1,n\}$. This possibility distribution is unique and is denoted $\pi^*$. Then the level of priority of rule $(A_i, B_i)$ is simply computed as $N^*(\neg A_i \vee B_i)$ (computed from $\pi^*$). Then given evidence A, and knowledge K, B is a plausible conclusion of A in the context B if and only if $N^*(B|A) > 0$, i.e. $\Pi^*(B \wedge A) > \Pi^*(\neg B \wedge A)$. This procedure suffers from the problem of blocking of property inheritance as shown in the following example.

$K = \{p \vdash \neg f, b \vdash f, p \vdash b, b \vdash l\}$ where p = penguin, b = bird, f = fly, l = legs. It is well-established that the rational inference method classifies the rules of K into 2 sets of rule: $\{b \vdash f, b \vdash l\}$ have lower priority than $\{p \vdash \neg f, p \vdash b\}$. It can be encoded in possibilistic logic as $N(\neg b \vee l) \geq \alpha$; $N(\neg b \vee f) \geq \alpha$; $N(\neg p \vee \neg f) \geq \beta$; $N(p \vee b) \geq \beta$, with $\beta > \alpha$.

The corresponding minimally specific ranking is such that $\Pi^*(p \wedge l) = \Pi^*(p \wedge \neg l)$, hence forbidding the conclusion that penguins have legs, despite the fact that the rule $b \vdash l$ is not involved in the conflict between penguins and birds with respect to flying. Several solutions have been suggested to solve this problem including maximal entropy rankings, lexicographic methods and others. Here we suggest that weak independence solves the problem.

Consider the graph induced by the rules of K.

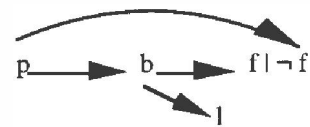

Any Bayesian-oriented AI researcher would suggest that l is conditionally independent of p in the context of birds (which is clearly not true for f). This is intuitive as well: If we learn that some bird is in fact a penguin, this does not influence our belief that it has legs. The conditional extension of weak independence reads

$$N(l|b) > 0 \text{ and } N(l|b \wedge p) > 0.$$

Here it leads to add the rule $p \wedge b \vdash l$ to K, i.e., to select another ranking of worlds that satisfies also $\Pi(l \wedge p \wedge b) > \Pi(\neg l \wedge p \wedge b)$ the level of priority of this rule will be the same as $p \vdash b$ and $p \vdash \neg f$. It is clear then that from p and $K \cup \{p \wedge b \vdash l\}$ one can deduce l plausibly.

There is no space to develop this point in detail here. However we plan to develop this methodology in the future (see, e.g. (Benferhat et al. 1994) for preliminary results). A first remark is that we do not use strong independence here. Strong independence would have two drawbacks

1) It would introduce equality constraints (here of the form $N(l|p \wedge b) = N(l|b)$) whose nature is different from that of the rules. As a consequence looking for the minimally specific possibility distribution that satisfies both rule-constraints and independence constraints may not lead to a unique solution. This is the problem already encountered by Goldzsmidt and Pearl (1992) with stratified rankings. The weak independence notion avoids this drawback.
2) It forbids the possibility of adding some contraposed rules since $N(l|p \wedge b) = N(l|b) > 0$ implies that $\Pi(p \wedge \neg l \wedge b) \geq \Pi(\neg p \wedge \neg l \wedge b)$, i.e., it is forbidden to claim that "birds without legs are not penguins" which seems to be a natural claim in the context of birds.

The idea of adding weak independence relationships to a rule base is to take advantage of the graphical structure of the knowledge base, as Bayesians do, and add just what is necessary. Part of the work is already done by the rational monotony property, i.e. $N(A|B) > 0$ and $N(\neg C|B) = 0$ does imply $N(A|B \wedge C) > 0$. However more conditional independence assertions are needed to overcome problems such as blocking of property inheritance. The problem is not to add too many assertions so as to avoid inconsistencies. Clearly we should stop imperatively once a *total* ordering of worlds is obtained. On the other hand the specification of conditional independence relation is extremely flexible and would enable to have tailored solutions to many inheritance problems. For instance if we add a bird that has no legs (n) to the above knowledge base, with rules saying that $n \vdash \neg l$, and $n \vdash b$, we can solve the problem by "reading on the graph" the proper conditional independence assertions while most other approaches would fail due to the presence of two conflicts.

However we cannot adapt the Bayesian methods readily for several reasons: here nodes of the graph are literals (not propositional variables), and cycles should be allowed (we must be able to say that "students are young" but "young people are not usually students"). Moreover there is no result that allow us to aggregate (via the min operation) a conditional possibility distribution into a global joint one (see, e.g., Fonck 1993). A third reason is that the weak independence relation is non-symmetric, i.e will not be a graphoid. Hence the mastering of weak conditional independence in the possibilistic setting for the purpose of handling exception-tolerant rule-bases is an open line of research, although a promising one.

## 8 CONCLUSION

This paper has provided a preliminary but systematic study of independence in the framework of possibility theory when conditioning is defined in an ordinal way (via the min-operation). The case where conditional possibility is defined as $\Pi(A \wedge C) = \Pi(C|A) \cdot \Pi(A)$ using product instead of min has been left for further research. It is also worth noticing that we have been working with events (or formulas) and not with variables (see (Studeny 1993) for an overview of the latter approach). It is well-known that in the probabilistic framework, the independence of A and B means, in terms of relative frequency, that the number of cases where A is true over the number of cases where A is false is left unchanged when B is known to be true. In the view of independence presented here, it can be checked that an analog property holds in terms of orderings: The possibilistic ordering between the interpretations with the greatest possibility which make A true and those which make A false is left unchanged when we restrict ourselves to interpretations where B is true. Besides, the transparency of probabilistic conditioning with respect to negation is closely related to the compositionality of probabilities with respect to negation. Similarly, the remarkable behavior of the possibilistic dependence and independence with respect to disjunction or conjunction stems from the fact that possibility measures are compositional with respect to disjunction, and necessity measures with respect to conjunctions.